# TRANSFERABLE ADVERSARIAL ATTACK ON IMAGE TAMPERING LOCALIZATION

*Yuqi Wang*[1, 2], *Gang Cao*[1, 2*], *Zijie Lou*[1, 2], *Haochen Zhu*[1, 2]

[1]State Key Laboratory of Media Convergence and Communication, Communication University of China, Beijing 100024, China
[2]School of Computer and Cyber Sciences, Communication University of China, Beijing 100024, China

**ABSTRACT**

It is significant to evaluate the security of existing digital image tampering localization algorithms in real-world applications. In this paper, we propose an adversarial attack scheme to reveal the reliability of such tampering localizers, which would be fooled and fail to predict altered regions correctly. Specifically, the adversarial examples based on optimization and gradient are implemented for white/black-box attacks. Correspondingly, the adversarial example is optimized via reverse gradient propagation, and the perturbation is added adaptively in the direction of gradient rising. The black-box attack is achieved by relying on the transferability of such adversarial examples to different localizers. Extensive evaluations verify that the proposed attack sharply reduces the localization accuracy while preserving high visual quality of the attacked images.

## 1. INTRODUCTION

Since digital image editing becomes easy, image authenticity is queried frequently. It is important to develop digital forensic techniques for detecting image forgeries. Many effective image tampering localization algorithms base on deep learning [1-8] have been proposed in recent years. Such algorithms effectively learn internal forensic traces from the training data. Specifically, the local consistency-based image splicing localizers, such as Noiseprint [6], EXIF-Net [7] and Forensic Similarity Graph [8], regard tampering localization as an anomaly detection problem. Such localizers rely on the extraction and consistency-checking of appropriate local features. There also exists another type of more effective localizers [1-5], which regard the localization as image semantic segmentation. The tampering localization map generated by encoder/decoder networks consists of pixel-level binary real/falsified labels.

It is significant to evaluate the security of such tampering localization algorithms against malicious attacks in real-world applications. Different from the image classification scenario, tampering localization involves the pixel-level prediction of tampering probability. As a result, it is necessary to specially address the adversarial attack on existing tampering localizers from the view of anti-forensics.

Previous attacks on image forensic algorithms are generally based on artificial features [9], Generative Adversarial Network (GAN) [10, 11] and adversarial examples [12, 13]. The artificial feature method is typically targeted to a specific forensic algorithm, and fails to address the deep learning-based tampering localizers. Xie *et al.* proposed a GAN-based method to attack the global manipulation detection schemes [10]. In the latest literature [11], forensic traces are synthesized by a two-phase GAN to deceive three local consistency-based image splicing localizers [6-8]. However, such a method can not be used for attacking the other major category of localizers, i.e., the segment-based ones [1-5]. Adversarial examples exploit the vulnerability of neural networks by adding minor perturbation to the inputs, resulting in some forensic errors [12]. In [14], the optimization-based adversarial example method is employed to attack the convolutional neural network (CNN)-based global manipulation detectors. Gradient of the output score function with respect to pixel values is explored in depth. The common gradient-based adversarial example algorithms including Fast Gradient Sign Method (FGSM) [15], Jacobian-based Saliency Map Attack (JSMA) [16] and Projected Gradient Descent (PGD) [17] have been used to attack the global manipulation detectors [12] and source camera identification models [13]. As far as we know, there are no prior works on attacking the tampering localizers via adversarial example.

To attenuate the deficiency of prior works, here we propose effective adversarial attacks on both the local consistency-based and segmentation-based tampering localizers. Specifically, two practical adversarial example methods are presented in a unified attack framework. In the optimization-based attack, the attacked image forgery is treated as the parameter to be optimized via Adam optimizer [18]. In the gradient-based attack, the invisible perturbation yielded by FGSM is added to the tampered image along gradient ascent direction. The transfer-based black-box attack is achieved by applying the generated adversarial example in white-box scenario to other localizers. Extensive evaluations verify the effectiveness of our proposed attack methods.

In the rest of this paper, the detailed attack scheme is proposed in Section 2. Performance testing experiments are given in Section 3, followed by the conclusion drawn in Section 4.

## 2. PROPOSED ADVERSARIAL ATTACKS

In this section, we first present the attack framework on tampering localizers. Then two specific attack methods are described in Subsections 2.2 and 2.3, respectively.

### 2.1. Attack framework on tampering localizers

Let the targeted tampering localization method to be attacked, namely victim localizer, be denoted by $y = f_\theta(x)$. Here, $x \in [0, 1]^{H \times W \times 3}$ denotes the normalized input image forgery with $H \times W$ pixels, and $y \in [0, 1]^{H \times W}$ is the pixel-wise prediction probability map. $\theta$ denotes model parameters. The pixel $x_{i,j,k}$ at the position $(i, j)$ with higher $y_{i,j}$ values towards 1 signifies the higher probability for a tampered pixel. Let $y^g \in \{0, 1\}^{H \times W}$ be the ground truth of the forged image $x$, where the values 1, 0 mark the tampered and pristine pixels, respectively. Let $x^*$ be the generated adversarial example image. The corresponding localization map $y^*$ predicted by the victim localizer $f_\theta$ is

$$y^* = f_\theta(x^*). \tag{1}$$

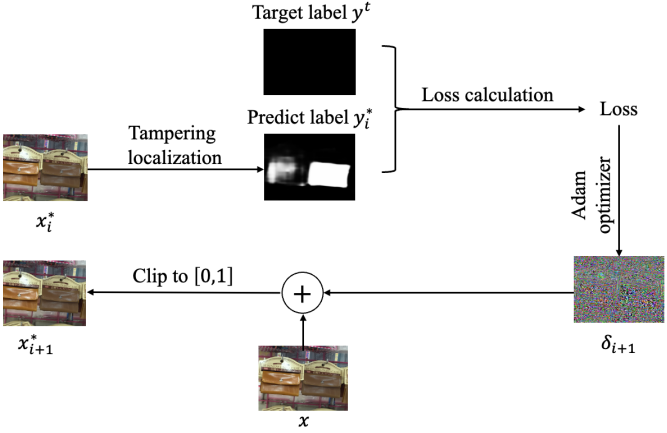

**Fig. 1.** The proposed optimization-based attack scheme against image tampering localization algorithms.

Generating adversarial examples can be formulated as finding an instance $x^* = x + \delta$, which satisfies the constrains as

$$\begin{cases} y^* \to y^t \\ D(x, x+\delta) \leq B \\ x + \delta \in [0,1]^{H \times W \times 3} \end{cases} \tag{2}$$

where $\delta$ is the perturbation quantity. $y^t$ is the target prediction probability map, and $y^* \to y^t$ denotes $y^*$ approaching $y^t$. The attack aims to find suitable $\delta$ that makes $y^*$ approach to $y^t$ while limiting the visual distortion $D(x, x+\delta)$ below a constant $B$. $D(\cdot,\cdot)$ is typically realized by $L_p$ norm. $x^*$ should still be a valid image.

Within the above framework, we propose two specific white-box attack methods for generating the adversarial example $x^*$, i.e., optimization-based and gradient-based attack. Note that the adversarial example yielded in such white-box attacks would be directly applied to the black-box attack against other localizers. The transferability of adversarial examples is exploited due to the limited knowledge of tampering localizers.

### 2.2. Optimization-based attack method

In this attack, the attacked forged image $x^*$ is regarded as the objective [19] to be optimized. In terms of Eq. (2), generating the adversarial example can be approximated as the following optimization problem:

$$\begin{aligned} & \underset{\delta}{\text{minimize}} && D(x, x+\delta) \\ & \text{such that} && y^* \to O^{H \times W} \\ & && x + \delta \in [0,1]^{H \times W \times 3}. \end{aligned} \tag{3}$$

It finds $\delta$ that minimizes $D$ and make $y^*$ tend to a zero matrix $y^t = O^{H \times W}$. Furthermore, Eq. (3) can be reformulated as

$$\begin{aligned} & \underset{\delta}{\text{minimize}} && \lambda \|\delta\|_2 + l(y^*, O^{H \times W}) \\ & \text{such that} && x + \delta \in [0,1]^{H \times W \times 3} \end{aligned} \tag{4}$$

where $l(\cdot,\cdot)$ is the binary cross-entropy (BCE) loss function that measures the distance between the prediction probability map $y^*$ and the target prediction probability map $O^{H \times W}$. $\lambda$ controls the proportion of perturbation, and the perturbation magnitude is measured by $L_2$ norm.

**Table 1.** Localization accuracy and visual quality comparison before and after the optimization-based attack on CASIAv1 dataset. The results of white-box attacks are underlined.

| Attack Method | | Before | Opt-OSN | Opt-MVSS | Opt-PSCC |
|---|---|---|---|---|---|
| F1 ($d \times 100$) | OSN[1] | .51 | .05 (90) | .23 (55) | .33 (35) |
| | MVSS[2] | .45 | .13 (71) | .03 (93) | .19 (58) |
| | PSCC[3] | .46 | .23 (50) | .23 (50) | .13 (72) |
| IoU ($d \times 100$) | OSN[1] | .47 | .04 (91) | .20 (57) | .29 (37) |
| | MVSS[2] | .40 | .10 (75) | .02 (95) | .15 (61) |
| | PSCC[3] | .41 | .20 (51) | .20 (51) | .11 (73) |
| PSNR (dB) | | — | 35.54 | 35.54 | 35.20 |
| SSIM | | — | 0.95 | 0.95 | 0.95 |

Detailed iterative process of the optimization-based attack is illustrated in Fig. 1. In each iteration, the adversarial example $x_i^*$ is input to the victim localizer for generating the prediction probability map $y_i^*$. Then the loss between $y_i^*$ and the target map $O^{H \times W}$ is calculated. $\delta_{i+1}$ is gained by solving the minimization problem described in Eq. (4) via Adam optimizer. Finally, the adversarial example image is updated by $x_{i+1}^* = x + \delta_{i+1}$ followed by clipping into the range [0, 1].

Note that the perturbation is added globally, since the local modification to tampered regions may still leave some new inconsistency. Besides, the loss value is also computed on global images. It aims to incur less difference between the tampered and unaltered regions.

### 2.3. Gradient-based attack method

Inspired by [12], the popular gradient-based adversarial example method FGSM [15] is used to attack tampering localizers. FGSM takes advantage of the linear approximation of victim localizers for fast generation. Adding a small perturbation in the gradient ascent direction can enlarge the loss value of the victim localizer dramatically. Generating the adversarial example $x^*$ via FGSM can be formulated as

$$x^* = clip\left(x + \varepsilon \cdot sign(\nabla_x l(y, y^g))\right) \tag{5}$$

where $l(y, y^g)$ is the loss function of the victim localizer at the training phase. By calculating gradient of the loss with respect to the input forgery image $x$, the perturbation is added in the direction of gradient rising denoted by $sign \nabla_x l$. The magnitude of perturbation is constrained by $\|\delta\|_\infty \leq \varepsilon$. Finally, to ensure the validity of generated adversarial examples, the clipping into the range [0, 1] is performed.

## 3. EXPERIMENTS

In this section, the performance evaluation experiments for the proposed attack methods are presented in detail.

**Table 2.** Localization accuracy and visual quality comparison before and after attacks on more other datasets and localizers. The results of white-box attacks are underlined.

| | Dataset | Columbia | | | Coverage | | | DSO | | | IMD | | |
|---|---|---|---|---|---|---|---|---|---|---|---|---|---|
| | Attack Method | **Before** | **Opt** | **FGSM** | **Before** | **Opt** | **FGSM** | **Before** | **Opt** | **FGSM** | **Before** | **Opt** | **FGSM** |
| F1 ($d \times 100$) | OSN[1] | .71 | <u>.12(83)</u> | <u>.26(64)</u> | .26 | <u>.11(58)</u> | <u>.13(52)</u> | .47 | <u>.01(98)</u> | <u>.00(100)</u> | .50 | <u>.04(92)</u> | <u>.00(100)</u> |
| | MVSS[2] | .64 | .55(14) | .53(17) | .45 | .21(54) | .24(48) | .30 | .17(43) | .21(29) | .27 | .11(59) | .14(48) |
| | PSCC[3] | .62 | .36(41) | .45(27) | .44 | .13(71) | .13(70) | .53 | .00(100) | .00(100) | .16 | .01(94) | .02(89) |
| | CAT[4] | .79 | .91(-15) | .92(-15) | .29 | .34(-17) | .33(-15) | .33 | .04(88) | .07(79) | .67 | .20(70) | .26(61) |
| | TruFor[5] | .81 | .73(10) | .71(12) | .53 | .34(35) | .36(32) | .90 | .35(62) | .42(53) | .72 | .43(41) | .47(35) |
| | Noiseprint[6] | .36 | .16(56) | .13(64) | .15 | .12(20) | .15(-3) | .29 | .04(86) | .05(84) | — | — | — |
| IoU ($d \times 100$) | OSN[1] | .61 | <u>.09(85)</u> | <u>.20(68)</u> | .18 | <u>.08(55)</u> | <u>.09(49)</u> | .34 | <u>.00(100)</u> | <u>.03(91)</u> | .40 | <u>.03(93)</u> | <u>.06(85)</u> |
| | MVSS[2] | .60 | .45(24) | .44(26) | .38 | .17(56) | .19(51) | .22 | .12(45) | .15(31) | .21 | .08(62) | .10(50) |
| | PSCC[3] | .48 | .27(44) | .35(28) | .34 | .11(67) | .11(69) | .42 | .00(100) | .00(100) | .13 | .01(92) | .01(90) |
| | CAT[4] | .75 | .88(-18) | .89(-19) | .23 | .26(-13) | .26(-12) | .28 | .03(89) | .04(85) | .59 | .15(75) | .20(65) |
| | TruFor[5] | .75 | .64(15) | .62(18) | .45 | .28(39) | .29(36) | .85 | .27(68) | .33(61) | .63 | .34(46) | .38(39) |
| | Noiseprint[6] | .26 | .09(65) | .08(71) | .09 | .07(21) | .09(-4) | .21 | .02(90) | .03(88) | — | — | — |
| | PSNR (dB) | — | 35.34 | 34.92 | — | 34.19 | 34.20 | — | 35.02 | 36.51 | — | 35.00 | 36.90 |
| | SSIM | — | 0.87 | 0.85 | — | 0.94 | 0.94 | — | 0.87 | 0.91 | — | 0.89 | 0.94 |

### 3.1. Experimental setting

*3.1.1. Datasets and localization algorithms*
Test datasets include CASIAv1 [20], Columbia [21], Coverage [22], DSO [23] and IMD [24] with 920, 160, 100, 100 and 2010 forged images, respectively. Due to limited computing resources, we follow the prior work [4] to crop some oversized images to 1096 × 1440 pixels for preparing the test images.

The attack performance is tested against six state-of-the-art image tampering localization algorithms: OSN [1], MVSS [2], PSCC [3], CAT [4], TruFor [5] and Noiseprint [6]. The first three are used as victim localizers for white-box attack. The officially released models of such localizers are used to generate adversarial examples. Noiseprint can not work on the CASIAv1 and IMD images due to too uniform content or small resolution.

*3.1.2. Evaluation metrics*
The localization accuracy metrics, i.e., F1 and Intersection over Union (IoU) are widely adopted by the existing works on tampering localization. F1 is the harmonic average of precision and recall, and IoU measures the similarity between the predicted area and the ground truth. Such metric values before and after the attack, and their decrease rate ($d$) are shown to evaluate the performance of attack methods. Here, $d$ is defined as the ratio between the decrement value and the measurement before attacks. Meanwhile, Peak Signal-to-Noise Ratio (PSNR) and Structural Similarity (SSIM) are used to evaluate the visual quality of attacked images.

*3.1.3. Parameter Setting*
In the optimization-based attack, the adversarial example is initialized with the forged image, i.e., $x^* = x$, the Adam optimizer is implemented with a learning rate of 0.003, and the number of iterations is set to 30 epochs. The optimal $\lambda$ that achieves a good attack performance while maintaining a certain level of visual quality is searched through experiments. We select the parameters that make the attack most effective, while keeping the PSNR greater than 34 dB. In the gradient-based attack, we set the step size as 0.02, 0.02, 0.001, 0.01 and 0.01 for the five datasets respectively to achieve similar PSNR values.

### 3.2. Influence of victim localizer in white-box attack

First, we apply optimization-based attack to MVSS, OSN and PSCC on CASIAv1 dataset for choosing the best victim localizer in the white-box scenario. Table 1 shows the F1, IoU and decrease rate before and after optimization-based white-box attacks and their transferability. It is observed that optimization-based attack can significantly reduce image tampering localization accuracy. Moreover, the decrease rate can exceed 70% in the white-box scenario. In addition, optimization-based attack shows strong transferability. The localization accuracy of image forensic methods has also degraded and it has reduced by at least 35%. As can be seen from Table 1, the selection of victim localizer has no obvious impact on the attack effect in the white-box scenario. The adversarial examples generated against OSN and MVSS have equal attack performance, while generated against PSCC are slightly less effective. The white-box attack against PSCC only decrease the localization accuracy by about 70%. The same conclusion is found in the gradient-based attack, too. For the sake of consistency and without loss of generality, OSN is chose as the victim localizer in all the following experiments for white-box attack.

### 3.3. Transferability in black-box attack

To further demonstrate the transferability of our attack, we evaluate the effectiveness of optimization-based and gradient-based attacks on more other datasets and localizers. The attack performance on four different datasets is presented in Table 2. We find that the adversarial examples generated in the white-box scenario have a certain transferability. In the white-box scenarios, optimization-based and gradient-based attacks can reduce the accuracy of the victim localizer by at least 49%. As can be seen from the results on the DSO dataset, the white-box attack can reduce OSN accuracy by nearly 100%. In white-box scenario, the knowledge about the victim localizer can be fully accessed to help attack the target localizer. As a result, white-box attacks can significantly degrade the performance of the victim localizer. However, the attack performance is not as obviously in black-box scenarios when compared to white-box attacks. Take the test results on the Columbia dataset as an example, the adversarial examples transferred to Noiseprint have the best attack performance, the decrease rate reach to 60%. However, while transferred to CAT, the localization performance after attack is even better. The black-box attack only uses the adversarial examples generated in the white-box scenario. And the victim localizer is not similar to the target localizer. Therefore, the attack performance in the black-box scenario is not as well as white-box scenario. Additionally, owing to the significant disparity between CAT and OSN, combined with CAT's insensitivity to minor perturbations, the accuracy of CAT is even improved when adversarial example is transferred to this localizer. As for the gradient-based black-box attack against Noiseprint on Coverage dataset, possibly due to poor performance before attack, the accuracy after the attack is slightly improved. It can be concluded that in most cases, the adversarial examples generated for OSN based on optimization and gradient perform well in both white-box and black-box scenarios.

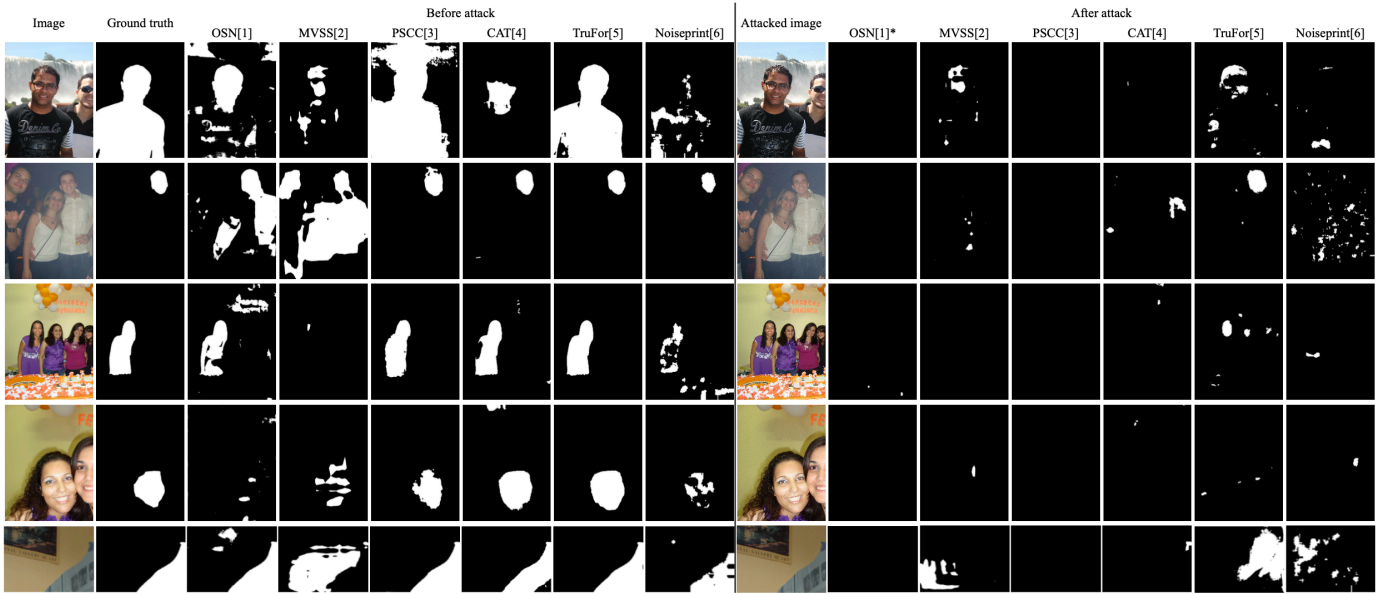

**Fig. 2.** Results of the optimization-based attack against different tampering localization algorithms on five example forged images. Here, * denotes the results of white-box attacks.

**Table 3.** Performance comparison with other attack methods on CASIAv1. The results of white-box attacks are underlined.

| | Attack Method | Before | JPEG [25] | Median [25] | JPEG-Median [25] | Opt | FGSM |
|---|---|---|---|---|---|---|---|
| F1 ($d \times 100$) | OSN[1] | .51 | .26 (48) | .37 (28) | .18 (64) | <u>.06 (88)</u> | .05 (90) |
| | MVSS[2] | .45 | .15 (68) | .39 (14) | .18 (59) | .12 (73) | .13 (71) |
| | PSCC[3] | .46 | .18 (62) | .26 (44) | .03 (93) | .24 (48) | .23 (50) |
| | CAT[4] | .72 | .29 (59) | .12 (83) | .17 (76) | .41 (43) | .40 (44) |
| | TruFor[5] | .69 | .57 (17) | .51 (27) | .44 (37) | .44 (36) | .49 (29) |
| IoU ($d \times 100$) | OSN[1] | .47 | .23 (51) | .31 (33) | .15 (67) | <u>.04 (91)</u> | .04 (91) |
| | MVSS[2] | .40 | .12 (71) | .33 (17) | .14 (64) | .09 (78) | .10 (75) |
| | PSCC[3] | .41 | .14 (66) | .19 (53) | .03 (94) | .21 (49) | .20 (51) |
| | CAT[4] | .64 | .24 (62) | .09 (86) | .13 (80) | .35 (45) | .34 (47) |
| | TruFor[5] | .63 | .50 (20) | .45 (28) | .38 (40) | .39 (38) | .43 (32) |
| PSNR (dB) | | — | 30.43 | 26.87 | 26.06 | 35.54 | 35.05 |
| SSIM | | — | 0.93 | 0.83 | 0.80 | 0.95 | 0.94 |

Qualitative evaluation results of the optimization-based attack against different localization algorithms are shown in Fig. 2. It can be observed that the adversarial perturbation is difficult to perceive. The predicted masks after attacks indicate that the localizers fail to locate the tampered regions accurately.

### 3.4. Performance comparison with other attacks

The code for GAN-based attack method [11] is not released, so we cannot compare our attack methods with it. Therefore, we compare proposed methods with common post-processing attacks on CASIAv1. JPEG compression [25] with the factor of 55, median filter [25] with the kernel 3×3 and JPEG compression after median filtering are tested. The comparison results are shown in Table 3. Compared with JPEG compression and median filtering attacks, adversarial attack can reduce the accuracy of the tampering localization algorithms while maintaining better visual quality. JPEG compression after median filtering has the best attack performance, except for TruFor. The localization accuracy of TruFor has only reduced by about 40%, all other localizers have reduced by more than 59%. However, such combined post-processing attack also leads to severely degraded images. The average PSNR of attacked images is only 26.06 dB. This indicates that the method sacrifices too much visual quality in order to improve attack performance.

## 4. CONCLUSION

In this work, we propose an effective adversarial attack scheme to evaluate the security of the state-of-the-art image tampering localization algorithms. The attack on tampering localizers is first formulated formally, then two specific adversarial example attack methods are presented under a unified attack framework. In both white and black-box scenarios, the accuracies of state-of-the-art tampering localizers are significantly reduced by our proposed attacks. Meanwhile, the adversarial example images enjoy good transferability and visual transparency. Our attack methods also outperform other existing attacks.